\documentclass[sigconf,nonacm,natbib=false]{acmart}

\makeatletter
\def\@ACM@checkaffil{
    \if@ACM@instpresent\else
    \ClassWarningNoLine{\@classname}{No institution present for an affiliation}%
    \fi
    \if@ACM@citypresent\else
    \ClassWarningNoLine{\@classname}{No city present for an affiliation}%
    \fi
    \if@ACM@countrypresent\else
        \ClassWarningNoLine{\@classname}{No country present for an affiliation}%
    \fi
}
\makeatother

\usepackage{cite}
 
\usepackage{amsmath,amssymb,amsfonts}
\usepackage[font=small,labelfont=bf]{caption, subcaption}
\usepackage[font=small,skip=0pt,labelfont=bf]{caption}
\usepackage[ruled,linesnumbered,vlined]{algorithm2e}
\usepackage{varioref}
\usepackage{hyperref}
\usepackage{colortbl}
\usepackage{graphicx}
\usepackage{textcomp}
\usepackage{xcolor}
\usepackage{supertabular}
\definecolor{Green}{rgb}{0,0.5,0}
\definecolor{Blue}{rgb}{0,0.5,1.0}
\usepackage{multirow}
\usepackage{makecell}
\usepackage{amsmath}
\usepackage{enumitem}

\AtBeginDocument{%
  \providecommand\BibTeX{{%
    \normalfont B\kern-0.5em{\scshape i\kern-0.25em }\kern-0.8em\TeX}}}

\setcopyright{none} 

\begin{document}



\title[]{ {ERIC: Estimating Rainfall with Commodity Doorbell Camera for Precision Residential Irrigation}}


\author{Tian Liu}
\affiliation{\small
    Computer Science \& Engineering\\
    \institution{Texas A\&M University}
}
\email{tian.liu@tamu.edu}

\author{Liuyi Jin}
\affiliation{\small
    Computer Science \& Engineering\\
    \institution{Texas A\&M University}
}
\email{liuyi@tamu.edu}

\author{Radu Stoleru}
\affiliation{\small
    Computer Science \& Engineering\\
    \institution{Texas A\&M University}
}
\email{stoleru@tamu.edu}

\author{Amran Haroon}
\affiliation{\small
    Computer Science \& Engineering\\
    \institution{Texas A\&M University}
}
\email{amran.haroon@tamu.edu}

\author{Charles Swanson}
\affiliation{\small
    Biological \& Agricultural Engineering\\
    \institution{Texas A\&M University}
}
\email{clswanson@tamu.edu}

\author{Kexin Feng}
\affiliation{\small
    Computer Science \& Engineering\\
    \institution{Texas A\&M University}
}
\email{kexin@tamu.edu}



\renewcommand{\shortauthors}{}

\begin{abstract}

Current state-of-the-art residential irrigation systems, such as WaterMyYard, rely on rainfall data from nearby weather stations to adjust irrigation amounts. However, the accuracy of rainfall data is compromised by the limited spatial resolution of rain gauges and the significant variability of hyperlocal rainfall, leading to substantial water waste. To improve irrigation efficiency, we developed a cost-effective irrigation system, dubbed ERIC, which employs machine learning models to estimate rainfall from commodity doorbell camera footage and optimizes irrigation schedules without human intervention. Specifically, we: a) designed novel visual and audio features with lightweight neural network models to infer rainfall from the camera at the edge, preserving user privacy; b) built a complete end-to-end irrigation system on Raspberry Pi 4, costing only \$75. We deployed the system across five locations (collecting over 750 hours of video) with varying backgrounds and light conditions. Comprehensive evaluation validates that ERIC achieves state-of-the-art rainfall estimation performance ($\sim$ 5mm/day), saving 9,112 gallons/month of water, translating to \$28.56/month in utility savings. Data and code are available at \url{https://github.com/LENSS/ERIC-BuildSys2024.git}.

\end{abstract}




\maketitle

\section{Introduction}
The U.S. Environmental Protection Agency (EPA) reports that landscape irrigation accounts for nearly one-third of all residential water use, amounting to over 9 billion gallons per day~\cite{EPA}.
Recent studies~\cite{EPA2} indicate that more than 50\% of water is wasted due to imprecise irrigation scheduling.
For instance, in East Texas, despite generally adequate rainfall, more than 90\% of residents over-irrigate their landscapes~\cite{pannkuk2015}. 
The significant waste of water motivates the design of more 
precise residential irrigation systems.

{
\begin{figure}[t]
\centering
\includegraphics[width=0.47\textwidth]{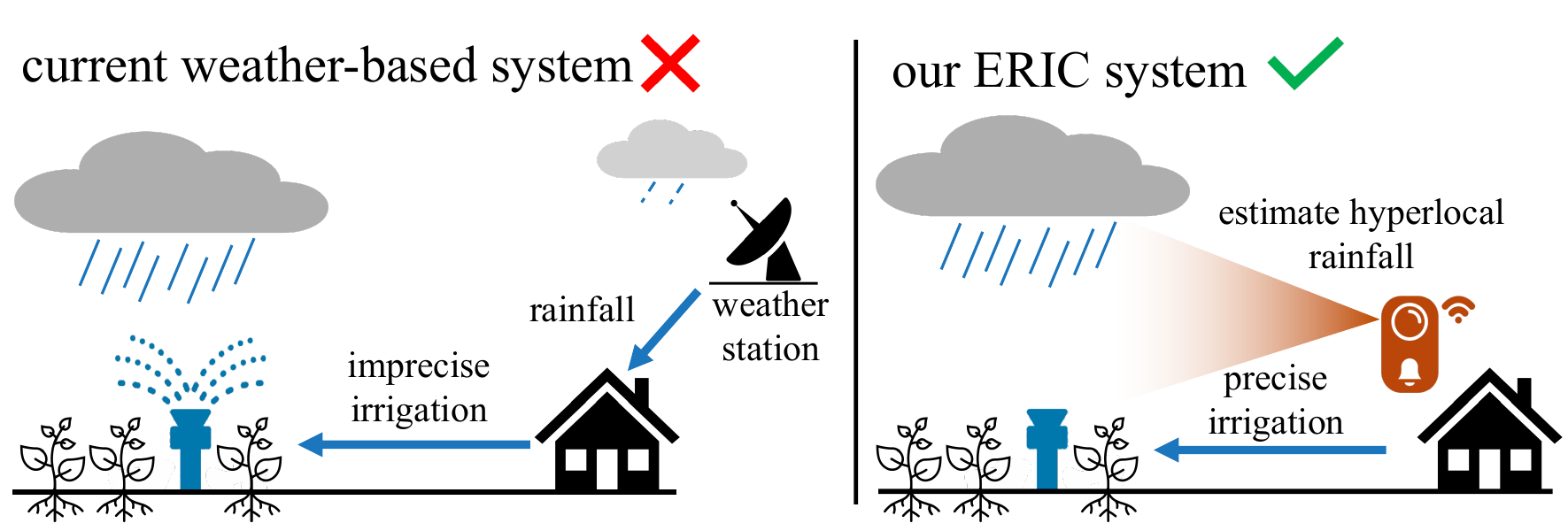}
\caption{
Comparison of the current weather-based irrigation system (left) with our ERIC system (right).
Current weather-based systems obtain rainfall data from nearby weather stations where the rainfall intensity can differ significantly from the hyperlocal rainfall at the residential sites, 
resulting in 
a significant waste of irrigation water. 
Instead, our ERIC system obtains accurate hyperlocal rainfall estimation from a doorbell camera, significantly improving irrigation precision.
}
\label{fig:illustration}
\end{figure}
}

{
\begin{figure}[t]
\centering
\includegraphics[width=0.49\textwidth, clip=true,trim = 5mm 0mm 0mm 0mm]{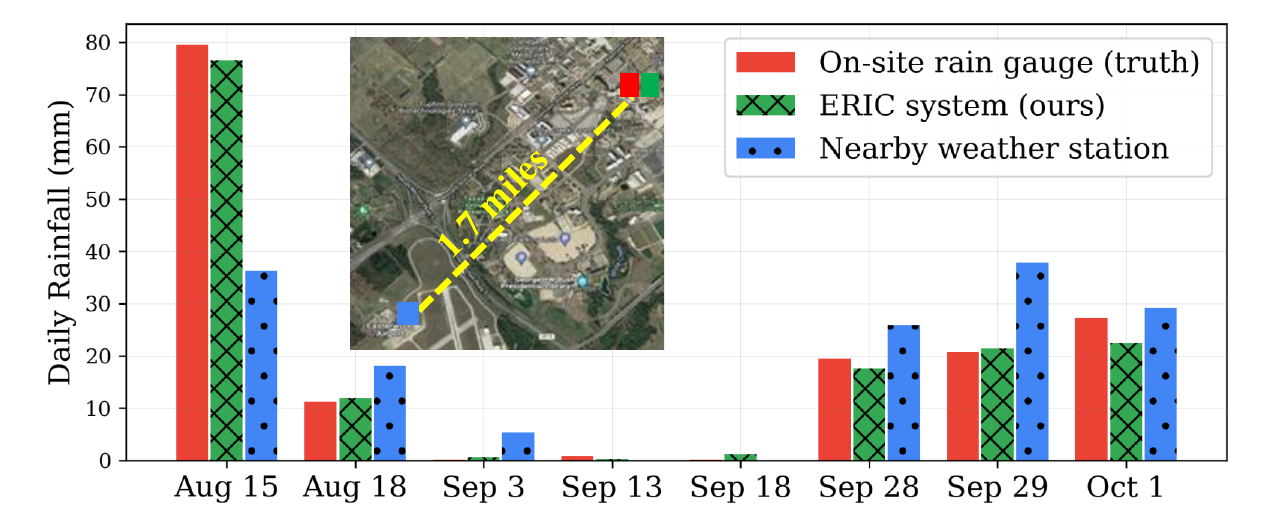}
\caption{
Our field experiment shows that rainfall measurements from a \textcolor{blue}{nearby weather station} that is only 1.7 miles away can differ as much as 54\% (43 mm on Aug 15) from the \textcolor{red}{true hyperlocal rainfall} as measured by a rain gauge at the residential site. However, our \textcolor{Green}{ERIC system} estimates hyperlocal rainfall accurately, saving over \$9,000 gallons of water for August 2021 (cf. Fig.~\ref{fig:water_saving}).
}
\label{fig:hyperlocal_rain}
\end{figure}
}

Traditional residential irrigation systems activate sprinklers on a fixed schedule, disregarding factors such as precipitation, solar radiation, plant types, and soil conditions, resulting in substantial water waste. Recent smart irrigation systems~\cite{EPAsmart} strive to deliver precise water by deploying soil sensors to monitor soil moisture (sensor-based) or retrieving weather data from nearby weather stations to calculate the water balance of soil-plant system, i.e., weather-based (also known as evapotranspiration-based or ET-based). However, limitations remain with both methods.

{\bf Sensor-based method is limited by high deployment costs.
} 
Previous studies~\cite{evett2008capacitance,teststd, UMsoil, peters2013practical, sharma2018methods} indicate that due to a limited sensing range (12 inches around the probe), multiple sensors are needed for adequate coverage, resulting in installation costs exceeding \$1,000, including the data logger. 
Additionally, these sensors require frequent calibration to the soil salinity, temperature, etc., further increasing maintenance costs ~\cite{UMsoil, irmak2014principles, werner1992measuring}.
The complexity of interpreting sensor data also reduces the practicality of these systems. 
As a result, current industry has widely adopted the more cost-effective weather-based method, e.g., WaterMyYard in Texas~\cite{wmy}, CIMIS in California~\cite{CIMIS}, and many others~\cite{AZMET, TIMS, IFAS}. 
Aligning with current industry practice, our
paper focuses on improving the precision of current weather-based
irrigation systems.

{\bf Weather-based method is limited by the inaccurate rainfall data from nearby weather stations.} 
This method accounts for the water balance in the soil-plant system, considering both outgoing water (evaporation from soil and plant transpiration) and incoming water (rainfall and irrigation)~\cite{UMet, harivandi2009managing}.
Accurate rainfall measurements from nearby weather stations are crucial for this method to calculate the desired irrigation amount. However, due to the spatio-temporal variability of hyperlocal rainfall, rainfall measured at a nearby weather station can 
be highly inaccurate (see Fig.~\ref{fig:illustration}).
Our field experiments show that rainfall measured from a weather station just 1.7 miles away can differ by as much as 54\% from the actual hyperlocal rainfall, leading to substantial over- or under-irrigation (c.f. Fig.~\ref{fig:hyperlocal_rain}).

{\bf Onsite rain gauges are expensive, limiting scalability.}
To achieve more accurate hyperlocal rainfall, recent weather-based irrigation controllers integrate on-site rain gauges, costing over \$300 for consumer-grade and over \$1,000 for professional-grade models~\cite{ETcontrollers, singh2020efficient}.
Besides the initial installation expenses, regular maintenance and calibration further escalate the overall cost.
Less expensive rain gauges (under \$100) are available, but they compromise accuracy and reliability and require technical skills for integration that many homeowners lack.

{\bf Can we estimate rainfall from doorbell camera?}
In this paper, we propose improving the current weather-based irrigation system with a low-cost, accurate solution for hyperlocal rainfall measurement. Inspired by the ubiquitous doorbell cameras and the high spatio-temporal resolution of video data, we explore the research question: “Can we develop a weather-based irrigation system that estimates hyperlocal rainfall from existing commodity doorbell cameras without additional hardware deployment?” We identify several key technical challenges below.

\begin{figure}[t]
\centering
\includegraphics[width=0.47\textwidth]{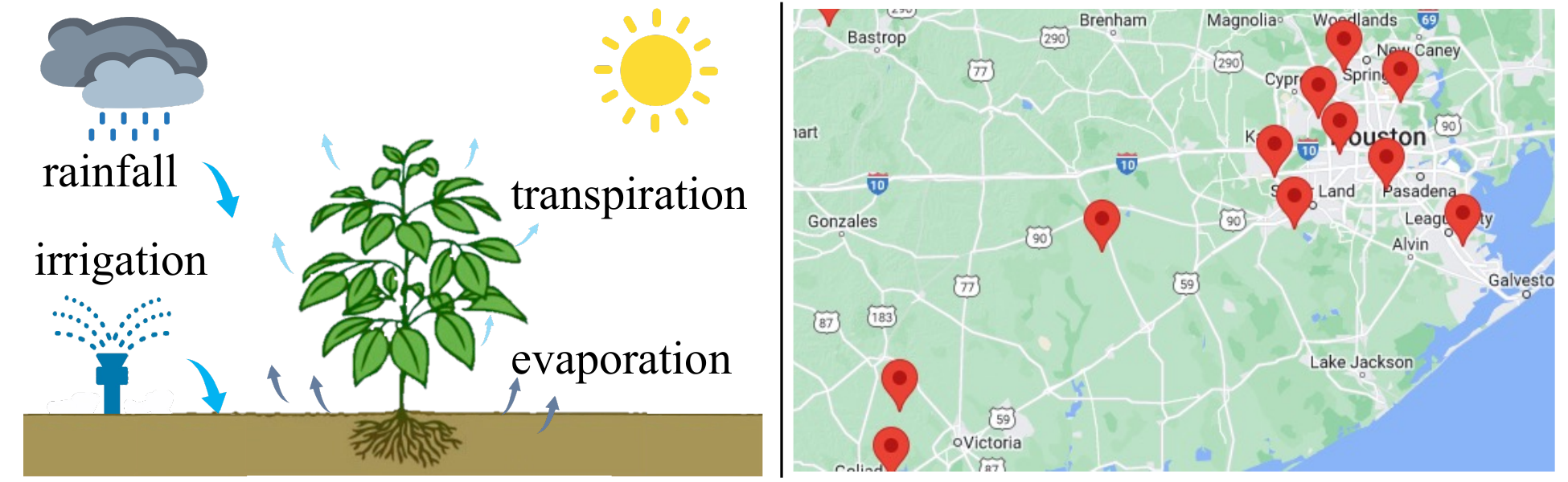}
\caption{
Left: an illustration of weather-based scheduling methods by considering the water balance between incoming water (rainfall, irrigation) and outgoing water (soil evaporation, plant transpiration). 
Right: a partial map of the industrial state-of-the-art weather-based program (WaterMyYard~\cite{wmy}) shows large spacing between weather stations, resulting in imprecise rainfall data for irrigation scheduling.
}
\label{fig:waterbalance}
\end{figure}
\normalsize

{\em Challenge 1: how to estimate rainfall from streaming video with high accuracy and low compute cost?} 
Prior rainfall estimation methods~\cite{garg2007vision, bossu2011rain, allamano2015toward, dong2017measurements, jiang2019advancing} employ computationally expensive rain steak extraction algorithms or Convolutional Neural Network (CNN) models that require high-end hardware (e.g. high-resolution cameras and GPU), which are unaffordable to common homeowners.

{\em Challenge 2: how to preserve user privacy as doorbell camera video is highly sensitive?}
Out of privacy concerns, homeowners are reluctant to upload doorbell camera recordings to the cloud for processing.
This challenges the real-time processing of streaming video data on resource-constrained edge devices to prevent data backlog.

{\em Challenge 3: how to evaluate our system in real-world deployment?}
Prior rainfall inference work evaluates with either images~\cite{yin2022estimating} or videos from traffic cameras~\cite{haurum2019raining}. Till now, there are no publicly available video datasets from residential environments. 
This lack of data necessitates creating new datasets and testing frameworks tailored to real-world residential settings to thoroughly assess the system's effectiveness.

{\bf Solutions and contributions.}
To address the above challenges, we present the design and implementation of ERIC system, the first precision residential irrigation system that harnesses existing commodity doorbell cameras and machine learning models to accurately estimate hyperlocal rainfall (cf. Fig.~\ref{fig:illustration}).
Specifically, we make the following contributions:

\textbf{1)} We developed lightweight neural network models based on our proposed reflection-based visual features and audio features for rainfall estimation. 
Our method achieves state-of-the-art performance with $\sim$ 5 mm/day error, using only low-cost commodity doorbell cameras.

\textbf{2)} We built an end-to-end irrigation system on a Raspberry Pi 4 device.
The simplicity of our models and pipelines enables ERIC system to process video data locally at the edge, preserving user privacy while ensuring low compute costs.

\textbf{3)} We deployed the ERIC system to five diverse real-world residential environments in two years. We collected over 750 hours of videos (including 150 hours of rain) with accurate rainfall ground truth from an onsite professional-grade rain gauge.

The following Section~\ref{sec:background} introduces related work. Section~\ref{sec:system} presents the system design and our models. 
Section~\ref{sec:implementation} shows our system implementation and 
Section~\ref{sec:evaluation} evaluates system performance in real-world deployments. 
Finally, conclusions and future work are discussed in Section~\ref{sec:conclusions}.

\section{Related Work}
\label{sec:background}

\subsection{Rainfall in Weather-based Systems}
\label{sec:water_balance}
The weather-based method (see Fig.~\ref{fig:waterbalance} left) calculates the irrigation requirement by considering the water balance between incoming water (rainfall and irrigation) and outgoing water ($ET\_loss$ via plant transpiration and soil evaporation)~\cite{harivandi2009managing, ETcontrollers, singh2020efficient}. 
Concretely, irrigation requirement $IR = ET\_loss - Rain$, where $ET\_loss$ is obtained from a nearby weather station based on measurements of solar radiation, temperature, wind, and humidity and adjusted for specific plant and soil types~\cite{wmy, UMet}. 
However, the rainfall measured from a nearby weather station could be highly inaccurate, due to the large spacing between weather stations (see Fig.~\ref{fig:waterbalance} right) and spatio-temporal variability of hyperlocal rainfall.
Such inaccurate rainfall measurements result in a significant waste of water, motivating us to leverage the ubiquitous doorbell cameras for hyperlocal rainfall estimation.

\subsection{Irrigation Optimization}
Previous work on irrigation optimization focuses on enhancing sensor-based methods. Winkler et al. ~\cite{winkler2016magic, winkler2018pics, winkler2019wisdom} developed a distributed sprinkler network with built-in soil moisture sensors, employing data-driven models to optimize irrigation.
However, the high deployment cost of sprinkler networks limits its scalability in residential irrigation.
Later work by Murthy et al.~\cite{murthy2019machine} improved weather-based methods by considering site-specific factors including soil and plant types, surface slope, etc., employing machine learning models with human feedback to prevent water run-off. 
Different from previous work, we focus on addressing the ever-neglected problem of inaccurate rainfall data from nearby weather stations. 
Our work aims to improve the irrigation efficiency of weather-based methods by providing accurate hyperlocal rainfall estimated from doorbell cameras.

\begin{figure}[t]
    \centering
    \includegraphics[width=0.49\textwidth]{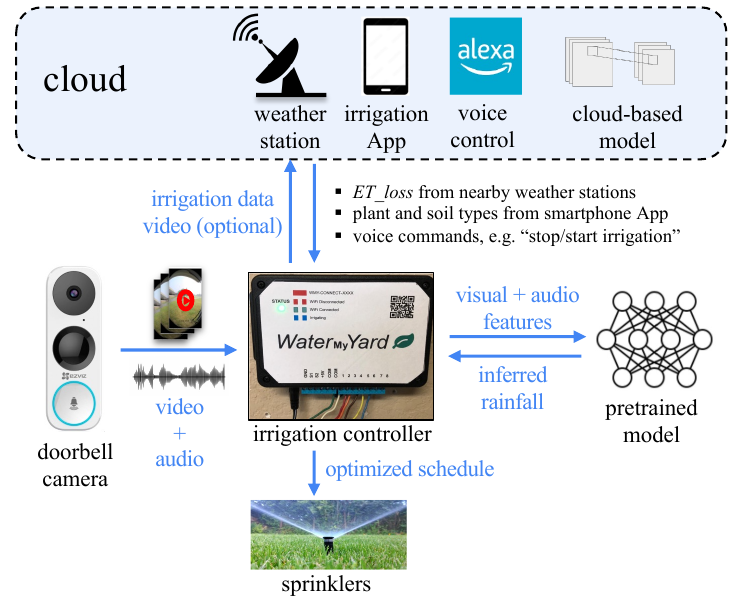}
    \caption{\small ERIC system architecture. ERIC system harnesses the existing doorbell camera to stream the video to the irrigation controller board and then leverages a lightweight pretrained model on the board to infer rainfall intensity at the edge using both visual and audio features. 
    Next, ERIC optimizes irrigation based on the estimated rainfall, $ET\_loss$ retrieved from nearby weather stations, and plant/soil types from user input on the smartphone App.
    Finally, it activates sprinklers according to the optimized schedule.
    ERIC also integrates Alexa for voice-controlled irrigation and allows users to upload videos to the cloud to use more powerful CNN models.     
    }
    \label{fig:rs_sys}
     \vspace{-7mm}
\end{figure}

\subsection{Rainfall Estimation from Camera}

Previous work on estimating rainfall from camera
can be categorized into extraction-based and deep learning-based methods.
Both methods have shown limitations.

{\bf Extraction-based methods are not robust to environmental factors}.
These methods~\cite{garg2007vision, bossu2011rain, allamano2015toward, dong2017measurements, jiang2019advancing} employ geometric and photometric models to first extract the foreground rain streak layer and then estimate rainfall intensity from the distribution of raindrop sizes.
The performance of these methods heavily relies on accurately capturing the fine-grained raindrop shapes, which not only requires 
expensive cameras (>\$2,000) and GPUs ~\cite{jiang2019advancing}
but also faces significant practical challenges. 
For example, the visibility of raindrops vanishes considerably with increasing distance to the camera. Additionally, the raindrop shapes can be largely distorted in a wide-angle camera. 
Moreover, 
none of these methods demonstrate feasibility under poor light conditions, e.g. nighttime with only a few street lights. 
All these challenging factors limit the practical effectiveness of extraction-based methods.

{\bf Deep learning-based methods suffer from large training costs, lacking rigorous evaluation.}
Recent works~\cite{haurum2019raining, avanzato2020cnn, yin2022estimating} leverage Convolutional Neural Networks (CNN) to detect or estimate the rainfall from images or traffic cameras. These models require a large training set along with significant training costs (long training time, expensive GPUs), which are not affordable to homeowners in residential environments. 
In addition, previous work evaluates their 
models with an oversimplified setup, i.e. testing on randomly split images~\cite{avanzato2020cnn, yin2022estimating} or only rainy videos with limited background~\cite{jiang2019advancing, haurum2019raining} rather than continuous video streams.
This is probably due to the lack of publicly available benchmarking video datasets.

{\bf Our novelties.}
In contrast to the aforementioned methods, our model: 1) estimates rainfall from intensity changes between video frames rather than raindrop shapes, demonstrating robustness to challenging environmental factors; 2) leverages lightweight neural networks, ensuring low training and inference costs; 3) evaluates with continuous streaming video (including both raining and non-raining) with diverse backgrounds, validating practical effectiveness.
We discuss more details 
in the following sections.

\section{System Design}
\label{sec:system}
In this section, we first present an overview of our ERIC system, and then introduce our machine learning models at the edge, followed by an extension to our cloud-based solution. Finally, we provide details for hardware design.

\subsection{ERIC System Overview}
    
Fig.~\ref{fig:rs_sys} presents an overview of our ERIC irrigation system. ERIC harnesses existing commodity doorbell cameras to stream video to the controller board, which then runs a pretrained machine learning model locally to estimate hyperlocal rainfall. Next, the controller board retrieves the latest $ET\_loss$ from a nearby weather station and calculates the irrigation requirement based on water balance (as discussed in Section~\ref{sec:water_balance}). Finally, the controller activates sprinklers based on the optimized irrigation schedules.

We highlight several advantages of our system:
1) {\bf high system efficiency}. Our model leverages a lightweight neural network model with our designed robust visual and audio features for rainfall estimation. Due to the simplicity of our model, it can process the streaming video data in real time without data backlog;
2) {\bf preserving user privacy}. Owning to the low compute cost of our model, all the sensitive video data can be processed locally on the controller board, which addresses homeowners' privacy concerns. 
For a complete system design, our ERIC does provide an extended option for users to upload their video data to the cloud, leveraging more powerful CNN models (e.g. ResNet~\cite{he2016deep}) for rainfall estimation, at the cost of privacy; 
3) {\bf low deployment costs.}
Our ERIC system runs efficiently on a Raspberry Pi 4 device, costing only \$75 in contrast to \$200-500 for the smart irrigation systems on the market (e.g. Rachio~\cite{Rachio}, Rainbird~\cite{RainBird}). In addition, ERIC schedules irrigation without human intervention, while the current industrial state-of-the-art weather-based program (WaterMyYard~\cite{wmy}) still requires homeowners to adjust the controllers weekly;
4) {\bf user-friendly interface}. We developed an Android App where the user can input plant and soil types, surface slopes, and other factors that are specific to their residential environment for optimal irrigation. The App also collects the irrigation history from the controller board and presents it to users. Moreover, we have integrated Alexa for voice-controlled irrigation, e.g. "start irrigation" or "stop irrigation" when noticing water runoff.

\begin{figure}[t]
    \centering
    \includegraphics[width=0.47\textwidth]{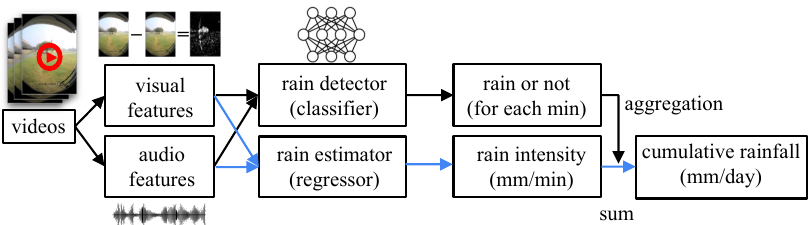}
    \caption{\small
    Our rainfall estimation workflow at the edge. Our pipeline extracts visual and audio features from the input video and feeds concatenated features to a rain detector and estimator to predict raining minutes and rainfall intensity, which are then aggregated into cumulative daily rainfall.
    }
    \label{fig:rain_infer_edge}
     \vspace{-3mm}
\end{figure}

\subsection{Rainfall Estimation at the Edge}

{\bf Rainfall estimation workflow.}
Given an input video file, our workflow (see Fig.~\ref{fig:rain_infer_edge}) first samples two adjacent frames every five seconds to calculate the visual features, which are then averaged per minute. Similarly, the audio features are extracted for each second and averaged for each minute. 
This sampling strategy is employed to reduce computation costs,
and the averaging mechanism is used to improve feature robustness against outliers (e.g. shaking camera, moving objects in the scene, etc.).
Next, the visual and audio features are concatenated and fed into a rain detector and a rain estimator model in parallel. The rain detector predicts whether each minute is raining or not, while the rain estimator predicts the rainfall intensity for each minute.
Finally, we aggregate the predictions by summing the rainfall intensity for raining minutes only, obtaining the cumulative daily rainfall. 
We emphasize the importance of our design, which includes both a rain detector and an estimator rather than solely relying on a rain estimator. This is because the rain estimator, a regression model, tends to predict small values (e.g. 0.01 mm/min) instead of zeros for non-raining minutes. Summing these small values would result in a significant deviation in cumulative daily rainfall. Our rain detector serves as a filter, effectively removing these errors.

{\bf Our intuition: estimating rainfall from raindrop reflections and sound.}
In contrast to previous work~\cite{garg2007vision, bossu2011rain, allamano2015toward, dong2017measurements, jiang2019advancing} which relies on accurate capture of raindrop shapes, our method estimates rainfall from the raindrop reflections and raining sound. 
Our key intuition is: {\em the reflections from the raindrops in air and splashes on the ground, and the volume of sounds of raindrops hitting surfaces strongly correlate to the rainfall intensity}.
In Fig.~\ref{fig:pseudo_rs}, we compare the RGB video frames along with their intensity changes under different rainfall intensities, light conditions, and backgrounds using our collected dataset. 
These intensity changes are calculated as the absolute differences between two adjacent frames after converting them into grayscale images, as defined by $\Delta I = |I_{n} - I_{n-1}|$~\cite{garg2007vision}. 
The intensity changes essentially capture the reflections from the fast-moving objects (e.g. raindrops, splashes) between adjacent video frames.
Interestingly, we make several important observations:

\begin{figure}[t]
    \centering
    \includegraphics[width=0.47\textwidth]{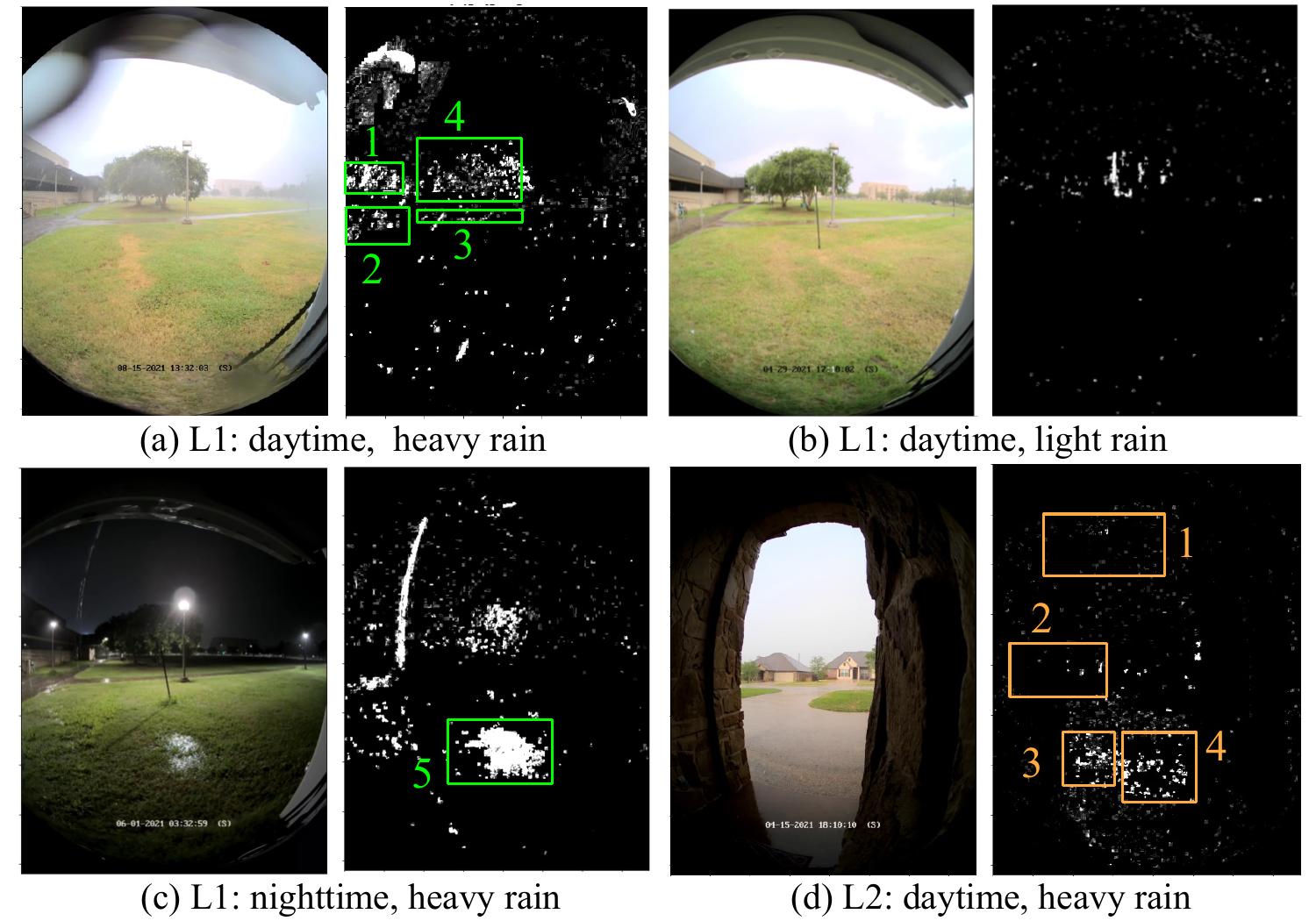}
    \caption{
    Comparison of RGB frames and intensity change maps under different rainfall intensities, light conditions, and environmental backgrounds. We mark the regions of interest (RoIs) that capture strong reflections from raindrops/splashes in boxes (\textcolor{Green}{green} shows manually identified RoI, \textcolor{orange}{orange} shows results of our proposed AutoRoI algorithm, cf. Fig.~\ref{fig:auto_roi}). The comparison shows that the significance of intensity changes strongly correlates to the rainfall intensity, demonstrating robustness against challenging light conditions and diverse backgrounds.
    }
    \label{fig:pseudo_rs}
     \vspace{-6mm}
\end{figure}

1) Fig.~\ref{fig:pseudo_rs} (a) shows that residual water on the camera lens can blur the captured frames while the raindrops at a further distance create a ``rain fog effect'', making it infeasible to extract the raindrop shapes accurately~\cite{jiang2019advancing}. This showcases the significant challenges of deploying previous extraction-based methods in practice. 

2) However, we find that the intensity changes easily capture the reflections from falling raindrops and splashes from the ground (see regions marked with green or yellow boxes), and strongly
correlate to the rainfall intensity.
For instance, heavy rain introduces much brighter and denser white dots on the intensity change map due to the larger and denser raindrops (Fig.~\ref{fig:pseudo_rs} (a)), while light rain yields much weaker and sparse white pixels (Fig.~\ref{fig:pseudo_rs} (b)). 

3) The intensity change maps work well under different light conditions (Fig.~\ref{fig:pseudo_rs} (c)) and diverse backgrounds (Fig.~\ref{fig:pseudo_rs} (d)), suggesting it provides a robust representation of rainfall intensity.

These observations motivate us to design reflection-based models to estimate rainfall. In addition, inspired by the repetitive sounds of raindrops, we explore combining audio information to improve the rainfall estimation performance.
Specifically, we identify several technical questions, which we address in the following.

\begin{figure}[t]
    \centering
    \includegraphics[width=0.48\textwidth]{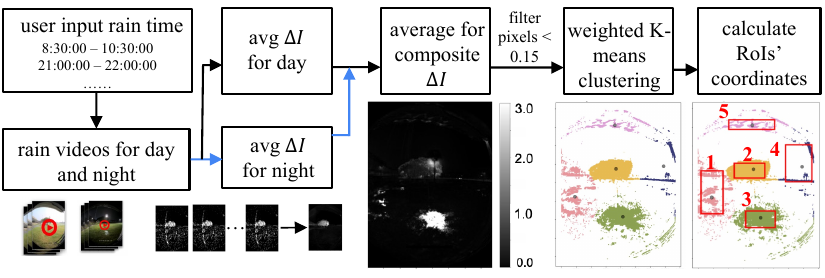}
    \caption{AutoRoI workflow. 
    Based on user's input on a few raining periods, AutoRoI fetches the raining videos and calculates the average intensity change maps for day and night, which are then averaged to obtain the composite map of $\Delta I$. Finally, the \textcolor{red}{RoIs} are obtained by K-means clustering.
    }
    \label{fig:auto_roi}
\end{figure}

{\bf Question 1: how to locate regions of interest that capture strong reflections?} 
To automatically locate the regions of interest (RoI) when deploying the ERIC system in diverse residential backgrounds, we propose the {\em AutoRoI} algorithm (see Fig.~\ref{fig:auto_roi}).
AutoRoI only requires the user to input the starting and ending times of a few raining periods. Next, it automatically fetches the corresponding videos to calculate the averaged intensity change maps for daytime and nighttime separately. These maps are then averaged to get the composite intensity change map, which emphasizes the regions with consistently strong reflections throughout the day. Then, a filter is applied to remove noisy weak pixels with $\Delta I < 0.15$ followed by a weighted K-means clustering. Finally, the bounding box for each cluster is defined by including 80\% bright pixels in the same cluster, which is achieved by taking the 10th and 90th percentiles of the sorted pixel coordinates.
Notably, AutoRoI runs only once in the training stage. The optimal number of clusters can be chosen using the validation set. During the inference, the model uses the obtained RoIs to calculate the visual features.
Our experiments in Section~\ref{sec:evaluation} validate the effectiveness of AutoRoI by comparing its performance with that of the handcrafted RoI.

\begin{figure}[t]
    \centering
    \includegraphics[width=0.48\textwidth]{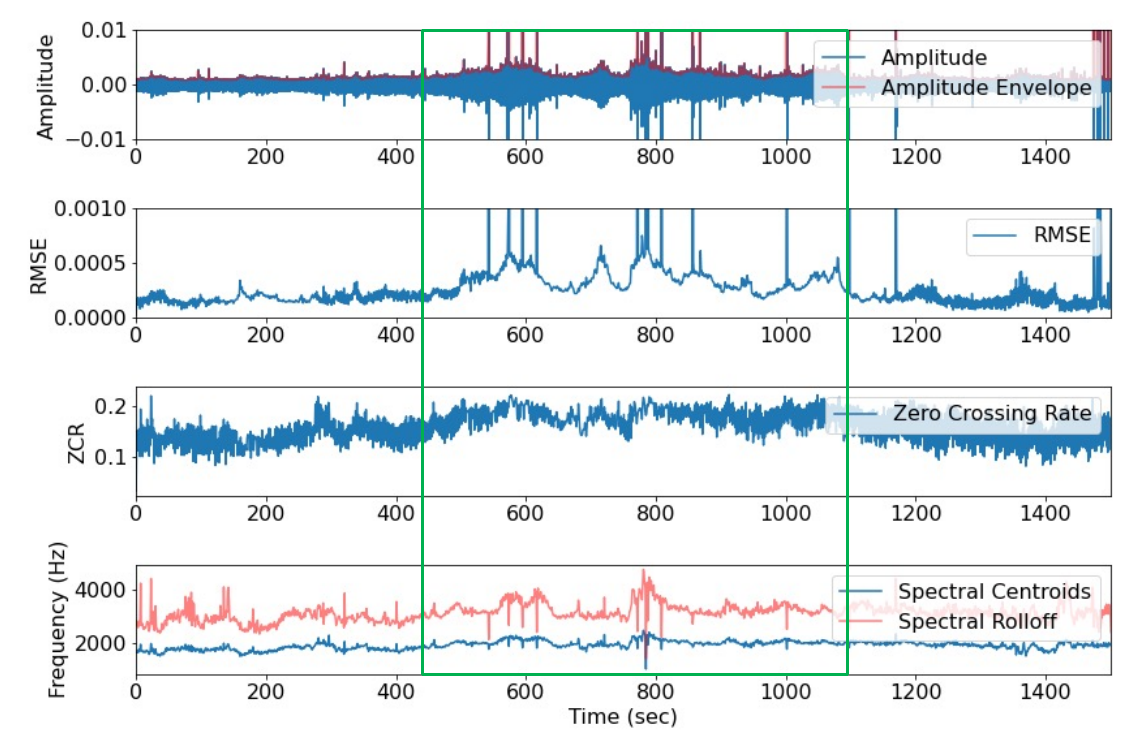}
    \caption{
    Plot of our selected audio features during rainfall. The raining period is marked in the \textcolor{Green}{green} box. Plots show that both amplitude-based features (AE, RMSE) and frequency-based features (ZCR, SC, SR) increase during the rain due to the distinct ``repetitive surface-hitting'' pattern of raindrops. The results
    clearly indicate a strong correlation between audio features and rainfall intensity.
    }
    \label{fig:audio_fea}
    \vspace{-4mm}
\end{figure}

{\bf Question 2: how to capture the visual and audio features?}
We designed robust visual and audio features to quantitatively measure the significance of intensity changes and the distinctiveness of raining sounds. 
Specifically, we design the following robust intensity-based visual features and calculate for each RoI.

(1) $max\_\Delta I$: the maximum intensity change. As shown in Fig.~\ref{fig:pseudo_rs}, heavier rain leads to higher intensity changes due to the larger size of raindrops.  

(2) $ brightness = \frac{N_{high}}{N_{total}}$: the fraction of high intensity-change pixels in a RoI, where $N_{total}$ is the total number of pixels and $N_{high}$ is the number of pixels with $\Delta I > threshold$, using the same threshold value of 3 as in \cite{garg2004detection}.

(3) $density = \frac{N_{bright}}{N_{total}}$: the percentage of bright pixels in a RoI. $N_{bright}$ is the number of pixels with $\Delta I > 0.1$. 
This feature captures the raindrops/splashes that are further away from the camera (e.g. RoI 1-3 in Fig.~\ref{fig:pseudo_rs}), which show much weaker reflections.

(4) $variability = sizeof(set(int(\Delta I_{i})))$, $i\in\{1... N_{total}\}$: the number of unique intensity change levels. This feature captures the light reflection and refraction from raindrops/splashes at various angles and distances, leading to large variations in pixel brightness.

In addition, to capture the distinct sound of rainfall (e.g. repetitively hitting a drum), we adopt the following low-level timbral features which are widely used in music genre classification or speech recognition~\cite{tzanetakis2002musical, baniya2013evaluation, 6973950}:
\textit{Amplitude Envelope (AE)}, \textit{Root-Mean-Square Envelope (RMSE)}, \textit{Zero Crossing Rate (ZCR)}, \textit{Spectral Centroid (SC)}, and \textit{Spectral Rolloff (SR)}. The \textit{AE, RMSE} are amplitude-based features reflecting the loudness of the sound, while \textit{ZCR, SC, SR} are frequency-based features representing the brightness of the sound.
Fig.~\ref{fig:audio_fea} showcases the increased audio features during the rain, suggesting the audio features embed useful information about rainfall intensity features.

\begin{figure}[t]
\centering
\includegraphics[width=0.48\textwidth]{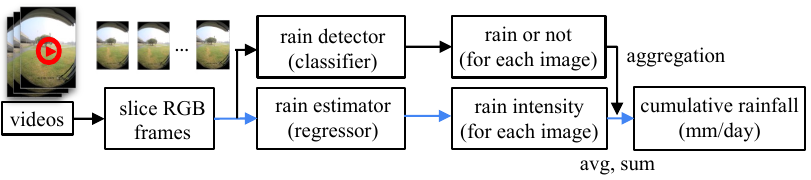}        
\caption{
Our rainfall estimation workflow in the cloud. 
The cloud-based solution leverages CNN models (ResNet18) to automatically extract features from sliced RGB frames for rainfall detection and estimation. The prediction results are then aggregated to obtain cumulative daily rainfall.}
\label{fig:rain_infer_cloud}
 \vspace{-6mm}
\end{figure}

{\bf Question 3: what machine learning models to use?} 
Unlike previous methods~\cite{haurum2019raining, avanzato2020cnn, yin2022estimating} that use computation-intensive CNN models, we design lightweight artificial neural network models (ANN) to ensure real-time inference and accommodate the limited computation power on the edge device (e.g. Raspberry Pi 4). Specifically, we implement a neural network model with only two dense layers, each containing 6 neurons. We use the {\em ReLU} activation function for the middle layers, {\em Sigmoid} function for the output layer of rain detector, and {\em Linear} function for the output layer of the rain estimator.
The simplicity of the model benefits from the effective visual and audio features we designed.
As a result, our model can process streaming video data in real time and conduct training and inference entirely at the edge.

\subsection{Extension: Rainfall Estimation in the Cloud}

Our model at the edge requires the user to input a few raining periods for AutoRoI calculation. 
To provide users with a fully automated option, we design a cloud-based solution that leverages CNN models for automatic feature extraction.
This option does require the user to upload their video data to the cloud, trading privacy for convenience. 
Similar to our edge solution, our cloud-based workflow (see Fig.~\ref{fig:rain_infer_cloud}) first slices video frames every five seconds and then passes them to the rain detector and estimator for rainfall prediction. The outputs are averaged per minute and then summed up for cumulative daily rainfall. We adopt the ResNet18 model~\cite{he2016deep} as the backbone for both the rain detector and estimator on the cloud.

\subsection{ERIC Irrigation Controller}
We build a smart irrigation controller board based on the Raspberry Pi 4 (8GB) device (see Fig.~\ref{fig:exp_setup} (b)), costing only \$75. 
We develop the controller based on \textit{OpenSprinkler}~\cite{OpenSprinkler}, with hardware and firmware enhancements. 
According to the irrigation schedule optimized by our ERIC, 
the controller turns on/off the irrigation valves through the Pi's GPIO pins, without using an additional microcontroller.
The firmware, written in C/C++, stores camera feeds, user data, and system configurations on the local storage.
The controller is powered by a 24V AC adapter which also powers the irrigation valves.
In addition, we developed a smartphone App through which users can input their plant/soil types for optimal scheduling, control the irrigation status (turn on/off), and view the irrigation history. 
Considering the convenience of voice assistants in various applications~\cite{jin2023emsassist, jin2023emsassistdemo, AlexaIrrigation2021, RainBirdAlexa2018, ahmed2021distributedirrigate}, we integrated the App with Alexa for voiced-controlled irrigation.

\section{System Implementation}
\label{sec:implementation}

In this section, we first introduce our implementation details of deploying the system to five real-world residential environments and then discuss the data collection and preprocessing.

\subsection{System Setup}
    \begin{figure}[t]
        \centering
        \includegraphics[width=0.47\textwidth]{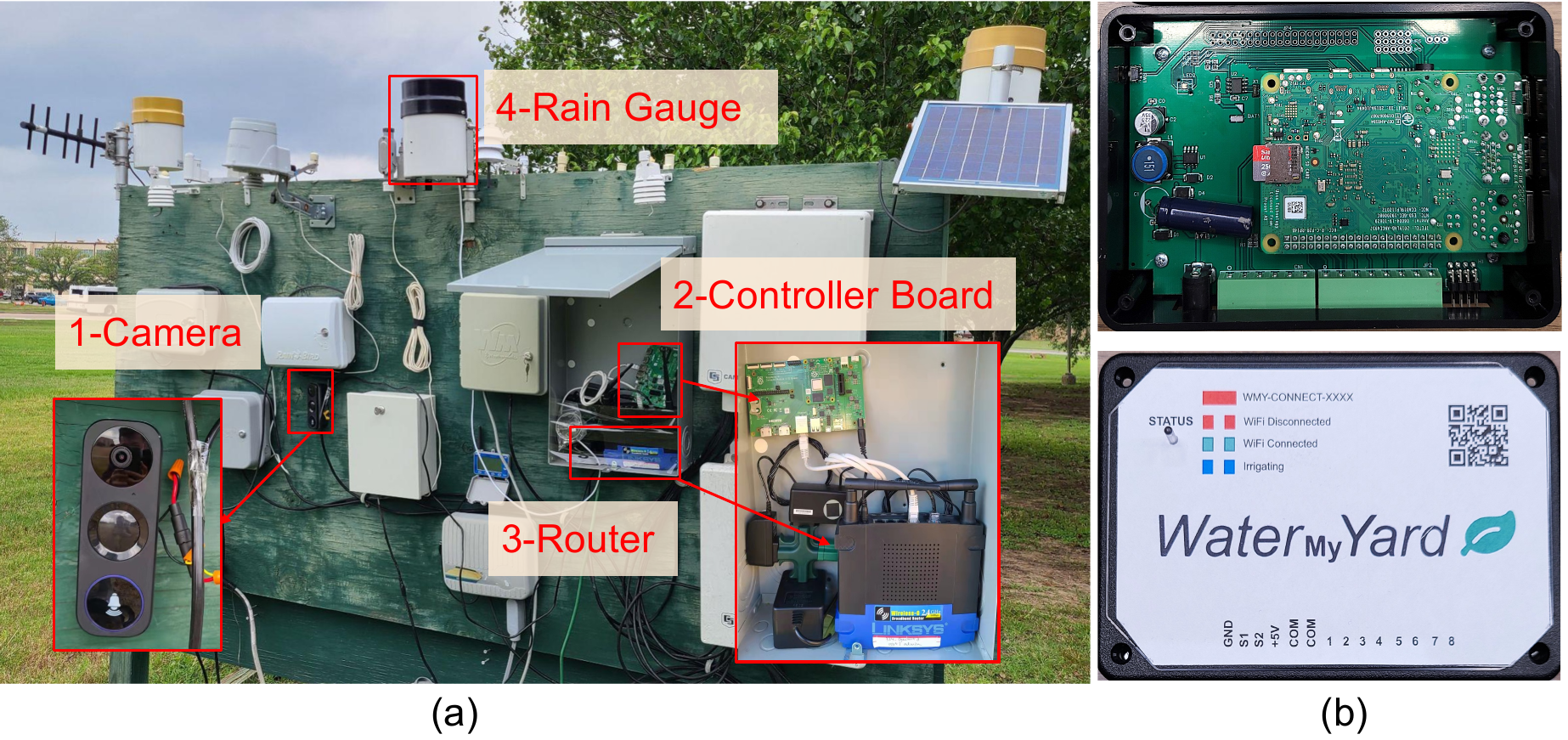}
        \caption{(a) Photo of ERIC system setup on the irrigation testing board at Location 1 (university campus). We highlight the four components: doorbell camera, controller board, router, and rain gauge. Note the rain gauge is only for measuring rainfall ground truth.
        (b) Our ERIC smart controller board developed based on Raspberry Pi 4 (8GB) device.} 
        \label{fig:exp_setup}
    \end{figure}

To test the practical effectiveness of ERIC, we deployed the system to five residential environments with varying camera types, backgrounds, and camera placements. These locations include 
a university campus (L1), the front door and backyard of three residential homes (L2, L3, L4, L5).

\begin{figure}[t]
    \centering
    \includegraphics[width=0.47\textwidth]{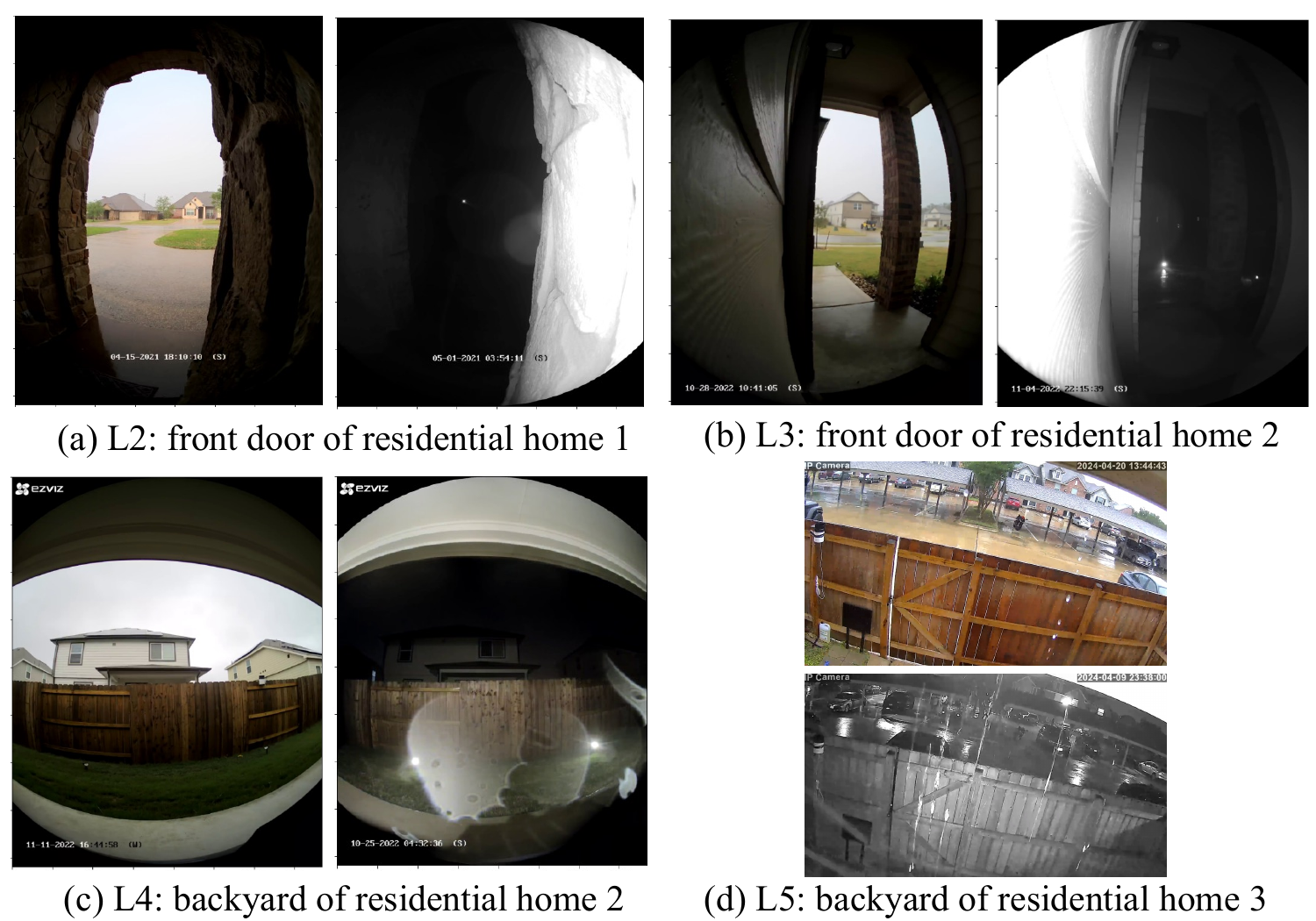}    
    \caption{
    Examples of camera footage (day and night) at five residential environments, showing the various camera placements, diverse backgrounds, and challenging light conditions at nighttime. See examples of L1 in Fig.~\ref{fig:pseudo_rs}.
    } 
    \label{fig:location2}
\end{figure}

{\bf Video data collection.}
Fig.~\ref{fig:exp_setup} (a) shows our system setup at L1. 
The camera (NSC-DB2) has a wide-angle lens and captures video in $1536\times2048$ resolution with 30 frames per second. 
It connects to the irrigation controller through the WiFi network provided by the router and streams real-time video data to the controller using the \textit{Shinobi} open-source software~\cite{shinobi}. The streamed video data are saved locally on the controller as 30-minute MP4 files.
We tested two types of cameras with different placements: a doorbell camera (NSC-DB2) installed on the door frames and a surveillance camera (Topodome) installed on the upper wall.
Fig.~\ref{fig:location2} illustrates examples of both daytime and nighttime video footage from each location.
Importantly, L2 and L3 present challenging light conditions at nighttime. 
Due to limited light sources, the camera can only capture views at a close distance.
Despite the poor light conditions, our edge model still demonstrates excellent performance in rainfall detection and estimation, highlighting the effectiveness of our robust visual and audio features (see Section~\ref{sec:evaluation}).

{\bf Rainfall ground-truth measurement.}
To measure the ground-truth hyperlocal rainfall,
we installed a professional-grade rain gauge right next to the camera at each location. 
We used a high-resolution tipping-bucket type rain gauge (HOBO RG3-M), costing \$800 with a data logger. The rain gauge collects the raindrops into a fixed-volume bucket and then tips to one side when it is full, triggering an electric signal that records a rainfall of 0.22 mm.

\subsection{Data Preprocessing}

We deployed the system to five residential locations, spanning over two years. We record the videos continuously but only save the videos of raining days due to storage limits. 
The resulting datasets amount to 750 hours (1 TB), including 150 hours of raining footage.
A detailed summary of collected data is provided in Table~\ref{tab:video_data}.

{\bf Video data splitting.}
To avoid information leakage from the random splitting of videos, we sequentially split the collected data into training, validation, and testing sets. 
For example, for the L1 dataset, we used videos of May, June, and July for training, August and early September for validation, and late September and October for testing.
We apply the sequential splitting to all datasets.

{
\setlength{\tabcolsep}{0.6em} 
\begin{table}[t]
  \renewcommand{\arraystretch}{.8}
    \caption{Overview of our collected five datasets with varying camera types, placements, and environmental backgrounds. 
    We calculate the percentage of raining and nighttime videos for each dataset.
    }
    \label{tab:video_data} 
    \scalebox{0.85}{
    \begin{tabular}{ccccccc}
    \toprule
    Location & Camera & Resolution & Videos & Raining & Nighttime \\
    \midrule
    \multirow{1}{*}{L1} &\multirow{4}{*}{NSC-DB2} &\multirow{4}{*}{1536 $\times$ 2048} & 232 hrs & 33.78\% & 25.32\%\\

    \multirow{1}{*}{L2} & & & 66 hrs   & 26.49\% & 22.09\%\\
    \multirow{1}{*}{L3} & &  & 186 hrs  & 6.67\% & 49.33\%\\

    \multirow{1}{*}{L4} & & & 158 hrs   & 7.64\% & 47.39\%\\   
    \midrule
    \multirow{1}{*}{L5} &\multirow{1}{*}{Topodome} &\multirow{1}{*}{ 1920 $\times$ 1080 }& 107 hrs   & 29.51\% & 43.72\%\\
    
    \bottomrule
    \end{tabular}}
\end{table}
}

{\bf Rainfall label preparation.}    
The raw data from the rain gauge is the cumulative rainfall (mm) recorded at each timestamp, with an interval of 0.22 mm. 
To train our machine learning models for rain detection (classification) and estimation (regression) tasks, we process the raw data to obtain the labels.
For the rain detection task, the labels are binary values for each minute of the videos. 
Due to the delayed activation of the tipping-bucket type rain gauge, i.e. the rain may have started much earlier before the bucket gets full and tips, 
we manually adjust the starting and ending time of each rain by checking the collected video footage. This process is necessary for obtaining accurate starting and ending time for light rains. For example, on Sep 18 (Fig.~\ref{fig:rain_esti_loc1}), the rain gauge only records one tipping. However, our manual correction helps recover the actual longer raining periods. The accurate labels improve the training of the model and ensure proper evaluation of the models.
For the rainfall estimation task, the labels are rainfall intensity (mm/minute). 
We obtain this by first interpolating the raw cumulative rainfall (mm) at each minute and then taking the difference between consecutive interpolated values.

\section{System Evaluation}
\label{sec:evaluation}

In this section, we evaluate our ERIC system using our collected datasets.
We first introduce evaluation metrics and baseline models and then provide a comprehensive analysis of ERIC's performance. 
For simplicity, in the following sections, we refer to our edge solution with ANN model as {\bf ERIC-edge} and our cloud-based solution using ResNet18 model as {\bf ERIC-cloud}.

\subsection{Evaluation Strategy}

{\bf Evaluation metrics.}
 For rain detection, we report the accuracy and F1 scores. Due to the significant imbalanced ratio of rain vs. not rain (see Table~\ref{tab:video_data}), we use the F1 score as the main evaluation metric because it considers both precision and recall.
 For rainfall estimation, previous work~\cite{haurum2019raining} uses mean absolute percentage error (MAPE), as defined by $MAPE = \frac{|R_{infer} - R_{true}| / R_{true}}{M}$, where M is the total number of rains. 
 However, we find that MAPE is strongly biased in light rains. For example, on Sep 18 in Fig.~\ref{fig:rain_esti_loc1}, the MAPE of our model is above 100\%, but the absolute error is only less than 1 mm. 
 To mitigate the biases, we designed two new metrics: total relative error, $TRE = \frac{|\sum_{i=1}^{N}R_{i, infer} - \sum_{i=1}^{N}R_{i, true}|}{\sum_{i=1}^{N}R_{i, true}}$, 
 and mean absolute daily error, $ MADE = \frac{\sum_{i=1}^{N}|R_{i, infer} - R_{i, true}|}{N}$, where N is the total number of days.
The TRE measures the relative error over the entire testing days, which reflects how the system performs over a longer period while MADE measures the absolute daily errors.

{\bf Compared models.}
We compare ERIC-edge and ERIC-cloud with the state-of-the-art extraction-based method by Jiang et al.~\cite{jiang2019advancing} and deep learning-based method, i.e. 3DCNN by ~\cite{haurum2019raining}.
Note that Jiang's method does not open-source their code so we can only compare with their reported MAPE score.
The 3DCNN model by ~\cite{haurum2019raining} was originally implemented for rain detection only. We adapt it for our rainfall estimation task by replacing the {\em Sigmoid} activation function of the output layer with {\em Linear} function.

{\bf Implementation details.}
We use {\em OpenCV} package to process visual features in videos and use {\em FFmpeg} and {\em librosa} to calculate the audio features. 
We implement models in ERIC-edge and ERIC-cloud using Tensorflow and PyTorch frameworks respectively.
We simulate the cloud by running experiments on a Linux server with 64 Xeon(R) Silver 4313 CPUs (2.4GHz) and 3 Nvidia A30 GPUs (24GB each).
We implement the 3DCNN model on our datasets using the provided code in ~\cite{haurum2019raining}.
For all models, we search hyperparameters on the validation set and report their performance on testing set.

{
\setlength{\tabcolsep}{0.18em} 
\begin{table}
\caption{
Comparison of detailed evaluation setup, models, and performance between our ERIC and previous state-of-the-art works. 
Our ERIC not only provides a more realistic evaluation with diverse backgrounds, longer videos (both rain and no rain), and challenging lighting conditions (nighttime) but also achieves much better rainfall estimation performance (i.e. lower MAPE) than previous works.
We calculate the MAPE averaged across five datasets for ERIC and 3DCNN and compare with the reported number by Jiang et al.~\cite{jiang2019advancing}.
}
\label{tab:comp_sota} 
\scalebox{0.88}{
\begin{tabular}{ccc|cc}
\toprule
&\makecell{Jiang et al.~\cite{jiang2019advancing}}  & 3DCNN~\cite{haurum2019raining} & \makecell{ERIC-edge\\(ours)} & \makecell{ERIC-cloud\\(ours)}\\
\midrule
\makecell{Camera model} &\makecell{EZVIZ C5Si}  &\makecell{AXIS M/Q-E}  & \multicolumn{2}{c}{NSC-DB2 / Topodome} \\
Camera cost & \$100 & \$300 & \multicolumn{2}{c}{\$30} \\
Background & Cropped roads & Cropped crossing & \multicolumn{2}{c}{Diverse residential} \\
Video size & 7 hrs &215 hrs & \multicolumn{2}{c}{750 hrs} \\
Rain condition &Rain only & Rain + no rain & \multicolumn{2}{c}{Rain + no rain}\\
Lightning & daytime only & daytime only & \multicolumn{2}{c}{daytime + nighttime} \\
Model &Decomposition &3DCNN & ANN &ResNet18 \\
\# of params &10 &0.45M &205 &11.7M \\
\midrule
MAPE &21.8\% &19.7\% &12.3\% &10.6\% \\
\bottomrule
\end{tabular}}
\end{table}
}

\subsection{Results}
We first compare our ERIC with previous works on evaluation setup, rainfall estimation performance, and compute cost, highlighting the practical effectiveness of ERIC. Then, we provide further analysis of ERIC's rainfall estimation performance, water savings, false positives and negatives, RoI selections, and input features.

\subsubsection{\bf Comparison with State-of-the-Arts.}
We compare the evaluation setup and model performance between our ERIC and previous SOTA by Jiang et al.~\cite{jiang2019advancing} and 3DCNN~\cite{haurum2019raining}.
As shown in Table~\ref{tab:comp_sota}, previous work only evaluates on videos with cropped backgrounds, limited duration containing either rains only or daytime only.
Instead, our work provides a more realistic evaluation setup by deploying our ERIC system in diverse residential backgrounds (see Fig.~\ref{fig:pseudo_rs}) and evaluating on longer videos with both rain and no rains, daytime and nighttime with challenging conditions.
Despite using a much cheaper camera and evaluating on a more challenging setup, our ERIC achieves lower MAPE than previous work, highlighting its practical effectiveness. 
Moreover, our collected video datasets serve as a realistic benchmark, providing opportunities for
more researchers to work on the rainfall estimation problem.

\subsubsection{\bf Comparison of Compute Costs.}
We further compare the compute costs between our ERIC with previous works. 
We measure their compute costs (time, memory, storage, etc.) for processing a 30-minute video file.
Table~\ref{tab:comp_cost} shows that our ERIC requires much less memory, GPU, and storage than previous work.
In fact, the previous method by Jiang et al.~\cite{jiang2019advancing} shows significant computation overhead, taking over three hours to process the file.
On the contrary, owing to our robust features and the simplicity of our ANN model, ERIC-edge runs efficiently on the Raspberry Pi 4 device, taking only 12 minutes to process a 30-minute video. 
The great efficiency enables real-time processing of sensitive video data at the edge, thereby preserving user privacy.
Our ERIC-cloud also shows better efficiency than 3DCNN~\cite{haurum2019raining}, due to its smaller model architecture.

{
\setlength{\tabcolsep}{0.38em} 
\begin{table}
\caption{
Comparison of compute cost between our ERIC and prior methods for processing a 30-minute video clip.
Our ERIC requires much less memory, GPU, storage costs than previous methods, enabling real-time processing of video data. ERIC-edge runs efficiently at the edge, preserves user privacy.
}
\label{tab:comp_cost} 
\scalebox{0.88}{
\begin{tabular}{ccc|cc}
\toprule
&\makecell{Jiang et al.~\cite{jiang2019advancing}}  & 3DCNN~\cite{haurum2019raining} & \makecell{ERIC-edge\\(ours)} & \makecell{ERIC-cloud\\(ours)}\\
\midrule

\makecell{Platform} &Workstation & Cloud & \makecell{Raspberry Pi 4} & Cloud \\





RAM &32 GB &10 GB &0.5 GB & 3 GB\\
GPU &12 GB &12 GB &0 GB & 5 GB \\
Storage &0.5 GB &4 GB & 0.5 GB & 1.5 GB\\
Time &3.3 hrs  &5 mins  &12 mins &1.5 mins\\
\midrule
real-time  &$\times$  &\checkmark  &\checkmark &\checkmark \\
\bottomrule
\end{tabular}}
\end{table}
}

\subsubsection{\bf Analysis of Rainfall Estimation Performance.}

\begin{figure}[t]
    \centering
    \includegraphics[width=0.47\textwidth, clip=true,trim = 0mm 43mm 0mm 0mm]{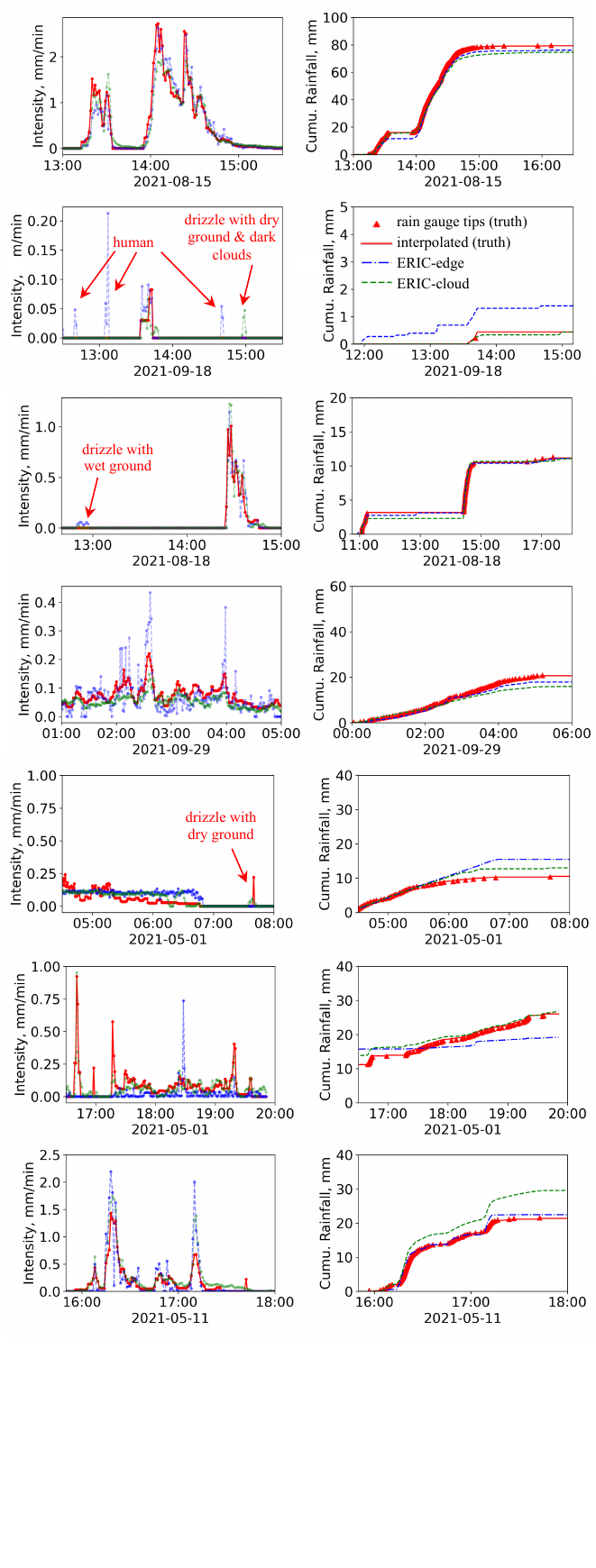}
    \caption{
    Comparison of rainfall intensity and cumulative rainfall between 
    \textcolor{blue}{ERIC-edge}, \textcolor{Green}{ERIC-cloud}, and an \textcolor{red}{on-site rain gauge} (ground truth). 
    We show examples from different datasets with diverse 
    residential 
    backgrounds, different rainfall intensities (from drizzles to heavy), and light conditions (daytime and nighttime).
    In all cases, ERIC-edge and ERIC-cloud predict close estimation to the ground truth, validating the effectiveness of our ERIC system.
    Causes of false positives are annotated in subplots.
    }
    \label{fig:rain_esti_loc1}
    \vspace{-6mm}
\end{figure}

We compare the rain detection and estimation performance of our ERIC-edge and ERIC-cloud with 3DCNN in Table~\ref{tab:rainfall_estimation}.
Results show that both our ERIC-edge and ERIC-cloud outperform 3DCNN significantly.
We hypothesize that this is because the 3DCNN model takes a sequence of frames (16 consecutive frames) as an input to capture the salient motions (falling raindrops in our case) across the frames~\cite{tran2015learning}, which could be sensitive to the unrelated moving objects in the scene.
Our analysis of 3DCNN's prediction by checking testing videos confirms our hypothesis. 
It shows that the 3DCNN model could be deceived by the moving tree leaves (windy sunny days) because the moving leaves create a similar motion as the falling raindrops. 
On the contrary, our ERIC-cloud considers the static information within a single RGB frame to identify the presence of rain, e.g. ``rain fogs'', wet ground, cloudy sky, darker background, etc. It shows more robust performance for both rain detection and rainfall estimation.
We show examples of detailed rainfall prediction plots for various locations in Fig.~\ref{fig:rain_esti_loc1}.

{
\setlength{\tabcolsep}{0.4em} 
\begin{table}[]
\caption{
Comparison of the rain detection and estimation performance between ERIC and previous 3DCNN model~\cite{haurum2019raining}.
We do not compare with Jiang et al.~\cite{jiang2019advancing} because they did not open-source their code.
We reimplement 3DCNN using the provided code in \cite{haurum2019raining} on our datasets. We highlight the {\bf best} score and \underline{underline} the second best.
Our ERIC significantly outperforms 3DCNN. ERIC-cloud performs slightly better than ERIC-edge.
}
\label{tab:rainfall_estimation} 
\scalebox{0.88}{
\begin{tabular}{cllllllll}
\toprule
Task & Method & Score & L1 & L2 & L3 & L4 & L5 & Avg \\
\midrule
\multirow{6}{*}{\makecell{Rain\\Detection}} & \multirow{2}{*}{3DCNN~\cite{haurum2019raining}} & Acc & 80.7 & 84.3 & 70.2 & 74.4 & 85.5 & 79.0 \\
 &  & F1 & 74.8 & 78.1 & 67.5 & 68.0 & 80.1 & 73.7 \\
 
 \cmidrule(lr){2-9}
 & \multirow{2}{*}{
 \makecell{ERIC-edge\\(ours)}} & Acc & 90.0 & 89.7 & 84.4 & 82.7 & 85.3 & \underline{86.4} \\
 &  & F1 & 81.9 & 80.6 & 81.0 & 80.6 & 81.4 & \underline{81.1} \\

  \cmidrule(lr){2-9}
 & \multirow{2}{*}{\makecell{ERIC-cloud\\(ours)}} & Acc & 88.1 & 91.1 & 85.0 & 85.4 & 84.4 & \textbf{86.8} \\
 &  & F1 & 82.3 & 86.3 & 82.6 & 82.4 & 85.6 & \textbf{83.8} \\

\midrule
\multirow{6}{*}{\makecell{Rain\\Estimation}} & \multirow{2}{*}{3DCNN~\cite{haurum2019raining}} & TRE & 0.20 & 0.41 & 0.25 & 0.17 & 0.23 & 0.25 \\
 &  & MADE & 4.42 & 14.5 & 13.8 & 8.65 & 9.60 & 10.19 \\
  \cmidrule(lr){2-9}
 & \multirow{2}{*}{\makecell{ERIC-edge\\(ours)}} & TRE & 0.10 & 0.23 & 0.10 & 0.13 & 0.12 & \underline{0.14} \\
 &  & MADE & 1.62 & 5.23 & 7.23 & 7.45 & 7.02 & \underline{5.71} \\
  \cmidrule(lr){2-9}
 & \multirow{2}{*}{\makecell{ERIC-cloud\\(ours)}} & TRE & 0.19 & 0.20 & 0.07 & 0.08 & 0.10 & \textbf{0.13} \\
 &  & MADE & 3.23 & 3.91 & 6.50 & 5.80 & 6.70 & \textbf{5.23} \\
\bottomrule
\end{tabular}
}
\end{table}
}

{
\begin{figure}[t]
\centering
\includegraphics[width=0.48\textwidth, clip=true,trim = 0mm 0mm 0mm 3mm]
{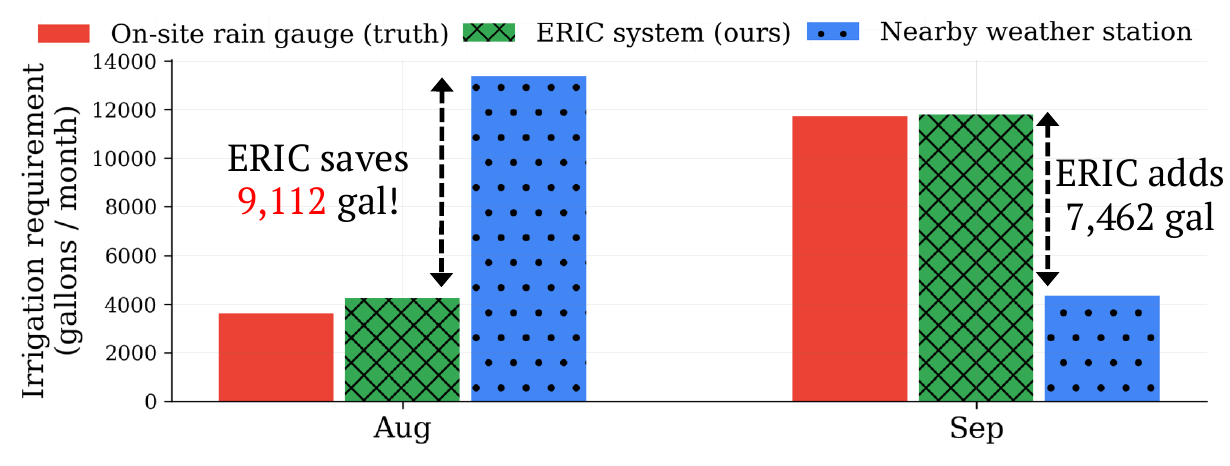}
\caption{
Comparison of irrigation requirement calculated using the rainfall obtained from an \textcolor{red}{on-site rain gauge}, \textcolor{Green}{our ERIC system}, and \textcolor{blue}{nearby weather station} (see Fig.\ref{fig:hyperlocal_rain}). 
ERIC saves over \$9,000 gallons of water for August, and provides the desired irrigation amount in September which avoids turf damage.
}
\label{fig:water_saving}
\end{figure}
}

\subsubsection{\bf Water Saving Analysis.}
Our ERIC system targets precision residential irrigation.
The key metric for evaluating its effectiveness is the reduction of both over-irrigation and under-irrigation. 
With more accurate hyperlocal rainfall (see Fig.~\ref{fig:hyperlocal_rain}), we analyze the effectiveness of ERIC by comparing
the calculated irrigation requirement ($IR$) for August and September using rainfall from an on-site rain gauge (ground truth), ERIC estimation, and nearby weather station. 
Fig.~\ref{fig:water_saving} shows that ERIC saves 9,112 gallons of water 
(\$28.56 utility savings) for August and provides the desired amount of irrigation water for September, avoiding potential turf damage due to under-irrigation.

\subsubsection{\bf Analysis of False Positives and Negatives.}
\begin{figure}[t]
    \centering
    \includegraphics[width=0.47\textwidth, clip=true,trim = 3mm 0mm 3mm 0mm]{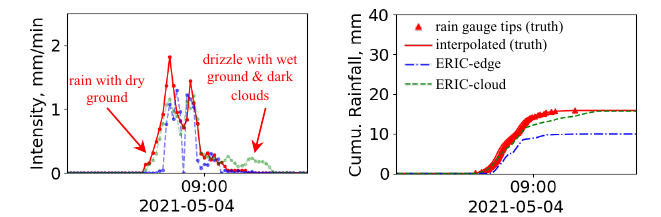}
    \caption{
    Examples of false positives or negatives. 
    The rainfall intensity plot (left) shows that ERIC-edge fails to detect the start of the rain, probably because the ground is still dry and there are no splashes yet. 
    However, ERIC-cloud accurately detects the rain, likely from the darker clouds.
    Moreover, ERIC-cloud detects the light drizzles at the end which does not even trigger the rain gauge. 
    }
    \label{fig:false_positive_plots}
\end{figure}

Fig.~\ref{fig:rain_esti_loc1}
shows that both ERIC-edge and ERIC-cloud accurately estimate the rainfall intensity and cumulative daily rainfall. 
Yet, some false positives and false negatives still exist. 
To understand why our models make these mistakes, we examined their false predictions by manually checking the recorded videos.
We annotate the causes on each subplot (cf. Fig.~\ref{fig:rain_esti_loc1}).
Interestingly, we made the following two observations:

1) {\bf ERIC can detect short drizzles that are not even recorded by the rain gauge!} 
For example, on May 4, 2021 at location 1, ERIC-cloud gives some false positive bumps (cf. Fig.~\ref{fig:false_positive_plots} left) after the end of recorded rain. 
After checking the video, we find that it was indeed a drizzle that was too light to even trigger the tipping bucket of the rain gauge.
This ``recovered true positive'' showcases the extraordinary capability of ERIC in detecting light rains, suggesting its potential to achieve even more accurate hyperlocal rainfall than the on-site rain gauge.

\begin{figure}[t]
    \centering
    \includegraphics[width=0.48\textwidth]{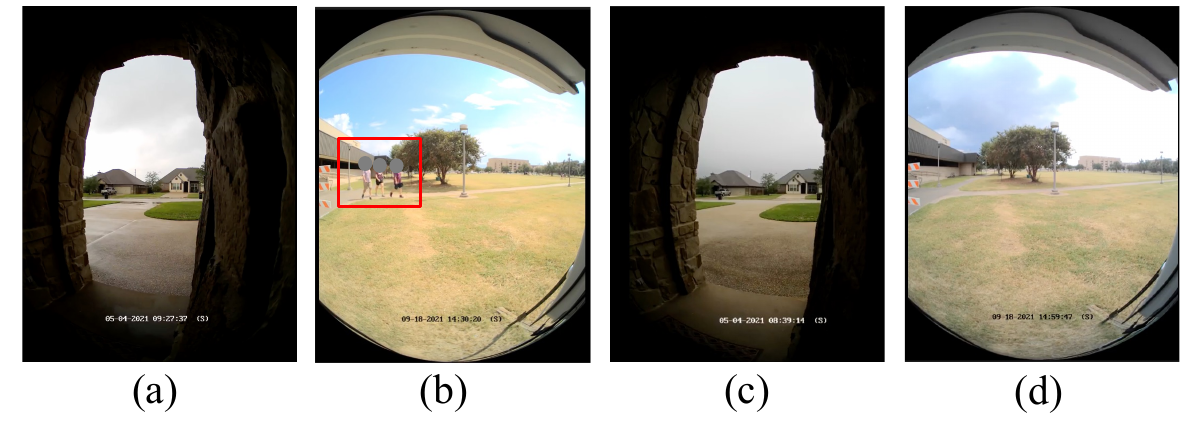}
    \caption{Example causes of false positives or negatives. 
    (a) Both ERIC-edge and ERIC-cloud detect the short drizzle that did not trigger the rain gauge. 
    (b) ERIC-edge gives false positives due to moving humans (marked in red box).
    (c-d) ERIC-edge fails to detect the rain as the ground is still dry, while ERIC-cloud correctly detects the rain likely from the darker clouds.
    }
    \label{fig:false_positives}
\end{figure}

2) {\bf ERIC-cloud is more robust than ERIC-edge.}
Because ERIC-edge detects rain from the reflections caused by the moving raindrops and splashes on the ground, it can be affected by the moving objects in the scene. 
An example is found on September 18, 2021 at location 1, where ERIC-edge predicts multiple false positive peaks due to the moving humans (cf. Fig.~\ref{fig:false_positives} (b)).
On the contrary, ERIC-cloud gives correct predictions. 
We hypothesize that this is because ERIC-cloud uses the ResNet model which detects the rainfall based on the static information from the entire image, e.g. wet ground, cloudy sky, darker backgrounds, etc. 
A supporting evidence is found on May 4, 2022, at location 2 (see Fig.~\ref{fig:false_positive_plots}), where ERIC-edge fails to detect the rain at the beginning because the ground was still dry and there were no splashes from the puddles on the ground. 
However, ERIC-cloud captures the start of the rain accurately, likely from the darker clouds. 
We show more examples of causes in Fig.~\ref{fig:false_positives}.

{
\setlength{\tabcolsep}{1.7em} 
\begin{table}[]
\centering
\caption{
Comparison of AutoRoI vs. ManualRoI on the performance of ERIC-edge on L1 dataset.
ManualRoI is slightly better than AutoRoI except on MADE. 
However, their performance gap is small (see Fig.~\ref{fig:compare_autoroi}).
We highlight the {\bf best} scores below.
}
\label{tab:compare_roi} 
\scalebox{0.88}{
\begin{tabular}{ccccc}
\toprule
 & Acc & F1 & TRE & MADE \\
\midrule
AutoRoI & 90.0 & 81.9 & 0.10 & \textbf{1.62} \\
ManualRoI & \textbf{91.2} & \textbf{84.2} & \textbf{0.07} & 1.96 \\
\bottomrule
\end{tabular}
}
\end{table}
}

\subsubsection{\bf Impact of RoI Selection (AutoRoI vs. Manual RoI)}
We evaluate our proposed AutoRoI algorithm by comparing its performance with manually annotated RoI.
We manually identify the regions with strong reflections and draw bounding boxes (cf. the green boxes in Fig.~\ref{fig:pseudo_rs}). Examples of AutoRoI are shown in Fig.~\ref{fig:auto_roi}.
Results in Table~\ref{tab:compare_roi} and Fig.\ref{fig:auto_roi} show only
a small performance gap between them, validating the effectiveness of our AutoRoI algorithm.

\begin{figure}[t]
    \centering
    \includegraphics[width=0.49\textwidth, clip=true, trim = 6mm 3mm 0mm 0mm]{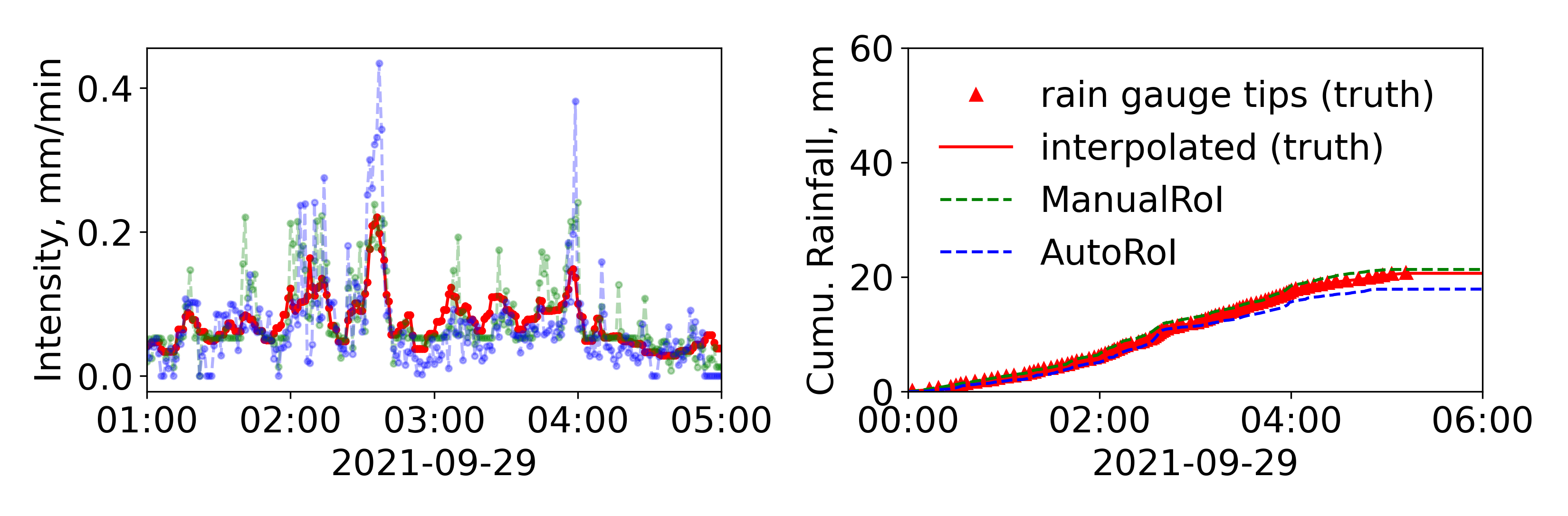}    
    \caption{ 
    While Manual RoI performs the best, AutoRoI performs closely as ManualRoI
    with a small difference in cumulative daily rainfall.
    Best viewed in color.
    }
    \label{fig:compare_autoroi}
\end{figure}

{
\setlength{\tabcolsep}{0.6em} 
\begin{table}[]
\centering
\caption{
Comparison of the rain detection and estimation performance of ERIC-edge on L1 dataset using different combinations of input features. 
Visual-only performs much better than audio-only, while adding visual and audio together obtains the best results. 
We highlight the {\bf best} scores below.
}
\label{tab:feature_modality} 
\scalebox{0.88}{
\begin{tabular}{llccc}
\toprule
 &  & Audio-only & Visual-only & Audio+Visual \\
 \midrule
\multirow{2}{*}{Rain detection} & Acc & 85.8 & 89.5 & \textbf{90.0} \\
& F1 & 75.1 & 80.0 & \textbf{81.9} \\
 \midrule
\multirow{2}{*}{Rain estimation} & TRE & 0.74 & 0.21 & \textbf{0.10} \\
 & MADE & 11.26 & 3.69 & \textbf{1.62} \\
 \bottomrule
\end{tabular}
}
\end{table}
}

\begin{figure}[t]
    \centering
    \includegraphics[width=0.49\textwidth, clip=true, trim = 6mm 3mm 0mm 0mm]{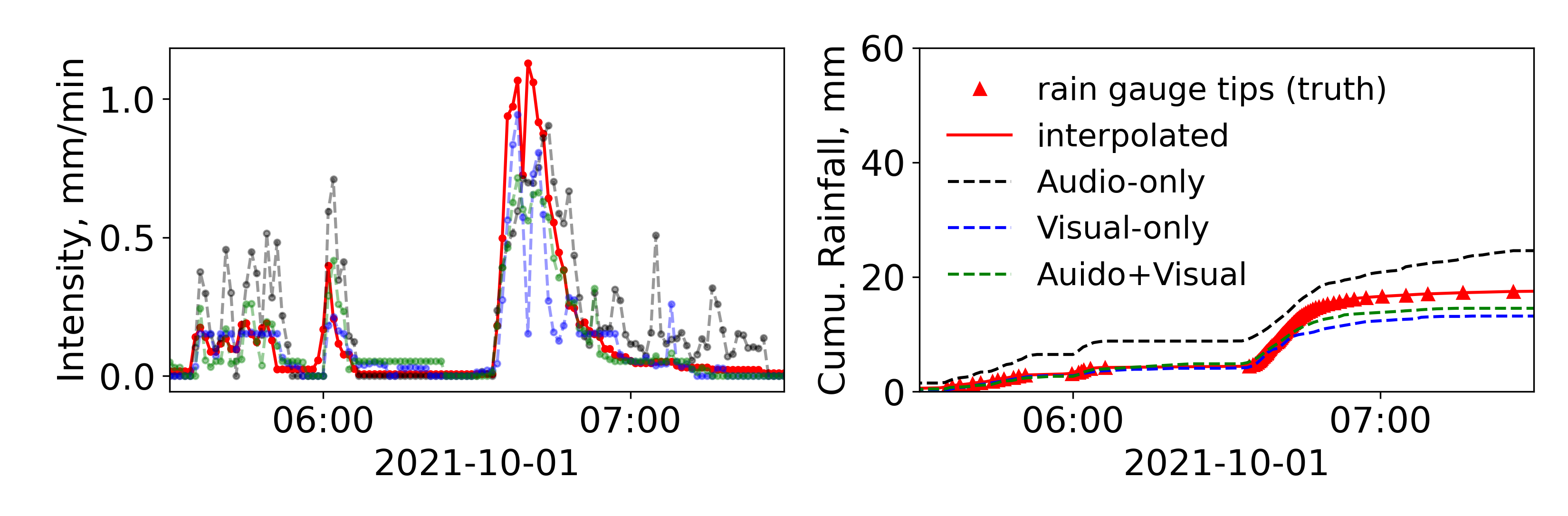}    
    \caption{ 
    Comparison of rainfall intensity and cumulative rainfall using different input feature modalities.
    Using audio features alone gives unstable spiky predictions on rainfall intensity (left), easily leading to over-estimation of cumulative rainfall (right). 
    However, using visual features provides better prediction, and using both 
    audio and visual features gives the best performance.
    Best viewed in color.
    }
    \label{fig:ANN_data_compare}
\end{figure}

\subsubsection{\bf Impact of Feature Modalities.}

ERIC-edge leverages both the visual and audio features for rain detection. 
We analyze the impact of each feature modality by comparing the performance of ERIC-edge with different input features:  audio-only, visual-only, and audio+visual. 
We share our interesting observations below.

{\bf Combining data of multi-modality performs the best.}
Table~\ref{tab:feature_modality} shows that using visual features achieves much better performance than using audio features while adding audio features still provides noticeable improvements. 
This is likely because audio features provide helpful information when the lightning conditions are poor.
This observation suggests that we should exploit all the useful information from data of different multi-modality for better rainfall estimation.

{\bf Audio features are more susceptible to environments.}
We compare the predicted rain intensity and cumulative rainfall in Fig.~\ref{fig:ANN_data_compare}. 
The plots show that using audio data tends to overestimate the rainfall intensity, as shown by the many spikes in Fig.~\ref{fig:ANN_data_compare} (left). 
This is probably because the audio data is more susceptible to different hardware (microphone) settings and background noises (e.g., car horns).
In contrast, using visual data yields a smoother intensity curve, validating the robustness of our visual features.

\begin{figure}[t]
    \centering
    \includegraphics[width=0.49\textwidth, clip=true, trim = 5mm 0mm 0mm 0mm]{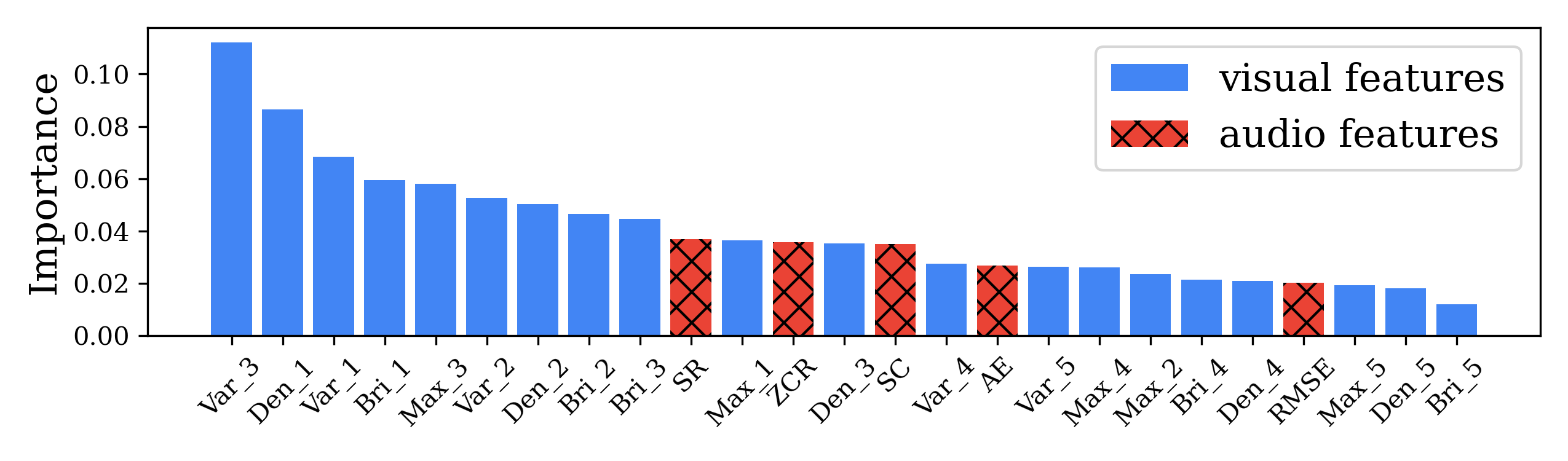}
    \caption{
    Ranked \textcolor{blue}{visual features} and \textcolor{red}{audio features} based on Gini importance for rain detection.
    Visual features generally show higher importance than audio features. Frequency-based audio features (e.g. SR, ZCR, SC) show higher importance than amplitude-based features (e.g. AE, RMSE). 
    }
    \label{fig:fea_importance}
\end{figure}

{\bf Visual feature shows higher importance than audio features.}
We rank the audio and visual feature importance in Fig.~\ref{fig:fea_importance}. 
Clearly, visual features show much higher importance than audio features, especially from the RoI box 1 and box 3 (see Fig.~\ref{fig:pseudo_rs}). 
This is likely because of the darker background (black buildings and windows) that leads to stronger reflections from the raindrops.
Additionally, the frequency-based audio features (\textit{SR}, \textit{ZCR}, \textit{SC}) show larger importance than those amplitude-based features (\textit{AE}, \textit{RMSE}). 
The reasons could be that the amplitude-based features are more easily affected by environmental noises like car horns or blowing wind, while the frequency-based features focus on capturing the ``repetitive surface-hitting'' pattern of raindrops.

\section{Conclusions and Future Work}
\label{sec:conclusions}
We present the design and implementation of ERIC, a cost-effective weather-based irrigation system that obtains accurate hyperlocal rainfall using existing doorbell cameras. By deploying ERIC to five diverse residential environments, our experiments highlight that ERIC achieves state-of-the-art rainfall estimation performance with significantly lower compute costs, saving over 9,000 gallons of irrigation water per month.
Our future work includes combining inputs from multiple surveillance cameras to further improve 
rainfall estimation performance
and automatically detecting water runoffs with cameras to improve irrigation precision. 
Moreover, considering the surprising zero-shot transfer capability of vision foundation models in recent works~\cite{clip, align, oquab2023dinov2, parashar2024neglected, liu2024few, abdullah2024ual}, we plan to explore adapting vision foundation models for rainfall detection and estimation.

\section{Acknowledgements}
\label{sec:acknowledgements}
This work was partially supported by Texas A\&M Water Seed Grant Initiative. 
We thank all anonymous reviewers for their constructive suggestions.

\bibliographystyle{unsrt}
\bibliography{references}


\end{document}